%% file: topk_filled.tex
\pdfoutput=1 
\PassOptionsToPackage{table}{xcolor} 
\documentclass[11pt]{article}

\usepackage[font=libertinus, citestyle=authoryear]{kurbanlab}

\input{affiliations}   

\usepackage{threeparttable}
\definecolor{good}{HTML}{B9E4BF}
\newcommand{\win}[1]{\cellcolor{good}{\bfseries\boldmath #1}}

\newcommand{\Vocab}{\mathcal{V}}
\newcommand{\Sset}{S}

\title{The Gradient That Top-\texorpdfstring{$k$}{k} Distillation Never Computes}
\RunningTitle{Tail mass in top-k distillation}

\Author{Rasul Khanbayov}{hbku}
\Author[corresponding=hkurban@hbku.edu.qa, orcid=0000-0003-3142-2866]{Hasan Kurban}{hbku}

\CodeURL{https://github.com/KurbanIntelligenceLab/top-k-distillation}

\begin{document}
\maketitle

\begin{abstract}
Many deployed distillation pipelines show the student a truncated
view of the teacher: keep the top-$k$ probabilities, renormalize, and minimize
a divergence on what remains. Renormalization is assumed to account for the
missing mass. We prove that it removes the visibility of tail mass, not the
freedom to move it. The renormalized objective depends only on log-odds among
the retained tokens, so at each position its gradient along the directions
that control tail mass is exactly zero; on a real teacher that component measures
$2$--$3\times10^{-15}$ of the gradient. The full-vocabulary gradient equals the
truncated gradient plus a residual weighted by the tail masses, and training
acts only on the first term. Without the
cross-entropy term, a top-$k$ student's tail mass drifts away from the
teacher's while a full-vocabulary control moves slightly toward it; with it, the gap
shrinks about fourfold but persists. On seven open-weight teachers, the retained
set determines the teacher's distribution only from $k=d+1$ ($d$ the hidden
dimension): recovery is exact from there on all seven, and commonly cached sizes
sit below it. In our
off-policy runs the practical effect is real but smaller than the choice of
teacher-student pair. Renormalization is a normalization, not a safeguard. The code is available at \url{https://github.com/KurbanIntelligenceLab/top-k-distillation}.
\end{abstract}

\section{Introduction}
\label{sec:intro}

Knowledge distillation minimizes a divergence between the student's and the
teacher's next-token distributions \citep{hinton2015distilling}. At vocabulary
sizes of $128{,}000$ to $262{,}000$, practitioners almost always truncate: keep
the teacher's top-$k$ probabilities, renormalize them, and compute the
divergence on the retained set. Reported $k$ spans top-$10$ \citep{sfd2026},
top-$16$ \citep{rethinkopd2026} and top-$32$ \citep{toolcall2026} through
offline caches of $k=100$ \citep{predistill2024,offlinetopk2026} to $k=128$
\citep{distillscaling2025}.
Whether this loses anything is contested. Several studies find truncation
harmless or beneficial
\citep{bild2024,deepkd2025,predistill2024,duet2026,offlinetopk2026}. Others
find it never reaches full-distribution distillation, degrades calibration
\citep{sparselogit2025}, costs $0.11$ nats at $k=128$
\citep{distillscaling2025}, and drops decision-critical tokens in tool use
while retaining $99.99$ percent of probability mass \citep{toolcall2026}. The
disagreement has been fought one training run at a time.

\begin{figure}[t]
\centering
\includegraphics[width=\linewidth]{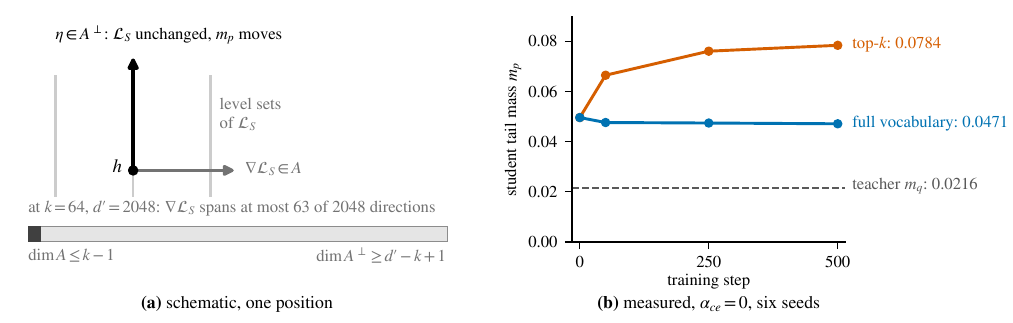}
\caption{The claim and its signature. \textbf{(a)} At one position the level
sets of $\mathcal{L}_\Sset$ are parallel to $A^{\perp}$, where $A$ is the span of
the retained rows' differences (\cref{thm:invariance}). So its gradient lies in
$A$ and is blind to the directions along which tail mass still varies
(\cref{eq:gradmass}); at $k=64$ and $d'=2048$, $A$ has dimension at most $63$
of $2048$. \textbf{(b)} With the cross-entropy term removed
(Llama-3.1-8B $\to$ Llama-3.2-1B, $k=256$, six seeds, seed spread below
$10^{-4}$), the top-$k$ student's tail mass moves away from the teacher's while
the full-vocabulary control's moves slightly toward it: mean $|m_p-m_q|$ at steps
$0,50,250,500$ is $0.0280\to0.0449\to0.0544\to0.0568$ against
$0.0280\to0.0260\to0.0258\to0.0255$.}
\label{fig:teaser}
\end{figure}

\paragraph{Visibility is not freedom.} Theory asks which divergence to minimize
and on which sequences \citep{agarwal2024gkd,gu2024minillm,ko2024distillm}, not
what the student is shown. The field's working assumption is that
renormalization accounts for the missing mass. \Citet{sfd2026} state it
explicitly: unnormalized top-$k$ training is clearly worse, and renormalization
closes the loophole by redistributing leaked mass. We show the assumption is
false for any retained set fixed by the teacher. Renormalization removes the \emph{visibility} of tail mass, not the
\emph{freedom} to move it. The renormalized objective depends only on
differences among retained logits. At any position, adding one constant to all
of them leaves it unchanged but changes how much probability sits outside the retained set. The
gradient in that direction is not small; it is zero (\cref{fig:same-loss},
\cref{fig:teaser}a).

\paragraph{Contributions.}
\begin{itemize}
\item \textbf{The truncated objective is blind to tail mass.} At each
position it is invariant on a subspace of dimension at least $d'-k+1$ of the
student's $d'$-dimensional hidden space, along which tail mass varies. With the cross-entropy term
removed, a top-$k$ student's tail mass moves away from its teacher's while a
full-vocabulary control moves slightly toward it (\cref{fig:teaser}b).
\item \textbf{An exact gradient decomposition.} The full-vocabulary gradient
equals the truncated gradient plus a term weighted by the tail masses. It holds summed over positions.
\item \textbf{The identifiability threshold, measured.} Below $k=d+1$, with $d$
the teacher's hidden dimension, the retained set does not determine the
teacher's distribution. On seven teachers recovery is exact from $k=d+1$; where
smaller $k$ can be measured, the error drops sharply there (\cref{tab:e2,tab:e2-curve}). Every operating point we found in the
literature sits below $d+1$ for its own teacher, by a factor of about $14$ to
$500$ (\cref{tab:cachesizes}).
\end{itemize}
We keep the practical reading narrow. Off-policy, the effect is smaller than
the teacher-student capacity gap (\cref{sec:exp}). At equal or smaller storage the
cache beats top-$1024$ on three of five teachers (\cref{tab:matched}). Our one
attempt to predict the published disagreement was falsified. Caching the
teacher's hidden state is deployed practice, so we claim no method
(\cref{sec:method}).

\section{Related work}
\label{sec:related}

Truncation is standard offline \citep{predistill2024,offlinetopk2026} and in
the top-$k$ variant of on-policy distillation, between sampled-token and
full-vocabulary supervision \citep{rethinkopd2026,sfd2026}. Each study above
measures the cost in its own setting. The most controlled
\citep{distillscaling2025} sweeps $k$ and reports the cost in nats, but not its
cause. No distillation study states what the retained set does
and does not determine; we supply that accounting rather than adjudicate between them
(\cref{app:relatedwork}). A separate literature studies which divergence to
minimize and on which sequences
\citep{kim2016seqkd,lin2020imitkd,agarwal2024gkd,gu2024minillm,wen2023fdivergence,ko2024distillm,ko2025distillm2}.
It now dominates post-training \citep{opdsurvey2026} and trades capability
against calibration \citep{illusioncertainty2026}. It asks what to
minimize; we ask what the student is shown.

Hidden-state caching is already deployed. Fused kernels
compute full-vocabulary objectives without materializing logits
\citep{wijmans2025cut,ligerkernel2024,decouplingkl2026,offlinetopk2026}.
\Citet{deepseekv4} cache last-layer teacher hidden states and reconstruct full
logits at training time. \Citet{kdflow2026} distill from pre-collected hidden
states. Both do so because transmitting top-$k$ logits breaks the equivalence
of the objective. Neither states how many observations identify the teacher,
what a truncated optimum leaves wrong, or how far exactness survives
quantization. The softmax bottleneck constrains logits to a $d$-dimensional
manifold \citep{yang2018breaking}. The same structure lets one recover a
model's hidden dimension and output layer from query access
\citep{finlayson2024logits,carlini2024stealing,stealgeom2026,seqlogits2025}. We
use it cooperatively. Feature distillation
\citep{romero2015fitnets,wang2020minilm} and the statistical view of
\citet{menon2021statistical} are discussed in \cref{app:relatedwork}.

\section{Setup and notation}
\label{sec:setup-notation}

Let $\Vocab$ be a vocabulary of size $V$. A teacher with hidden dimension $d$
produces, at a given position, a hidden state $g \in \R^d$ after its final
normalization, and logits
\begin{equation}
z = U g + b, \qquad U \in \R^{V \times d},\; b \in \R^{V},
\label{eq:logits}
\end{equation}
with unembedding $U$ and optional bias $b$; $q = \mathrm{softmax}(z)$. A
student with hidden state $h \in \R^{d'}$ and unembedding
$W \in \R^{V \times d'}$ produces $p = \mathrm{softmax}(Wh)$. We write
$u_v^\top$ and $w_v^\top$ for rows of $U$ and $W$; vectors and matrices are
set in plain italics, and calligraphic letters denote sets.

Fix a retained set $\Sset \subseteq \Vocab$ with $|\Sset| = k$, the
\emph{head}, and a reference token $v_0\in\Sset$; the complement of $\Sset$ is
the \emph{tail}. Truncation renormalizes both distributions on $\Sset$:
$\tilde q_v = q_v / \sum_{u\in\Sset} q_u$ and
$\tilde p_v = p_v / \sum_{u\in\Sset} p_u$ for $v \in \Sset$. The tail masses are
$m_q = 1-\sum_{v\in\Sset} q_v$ and $m_p = 1-\sum_{v\in\Sset} p_v$, and we call
$|m_p-m_q|$ the student's \emph{drift}. The full and truncated forward
objectives are
\begin{equation}
\mathcal{L}(h) = \KL(q \,\|\, p), \qquad
\mathcal{L}_\Sset(h) = \KL(\tilde q \,\|\, \tilde p).
\label{eq:objectives}
\end{equation}
Practical recipes add a cross-entropy term on the data with weight
$\alpha_{ce}$; our theorems concern $\mathcal{L}_\Sset$ itself. Both gradients
are differences of averaged student unembedding rows. The subscript says which
part (head or tail), the superscript which distribution supplies the weights:
\begin{equation}
\begin{aligned}
c_S^q &= \sum_{v \in \Sset} \tilde q_v w_v, &
c_T^q &= \frac{1}{m_q}\sum_{v \notin \Sset} q_v w_v,\\
c_S^p &= \sum_{v \in \Sset} \tilde p_v w_v, &
c_T^p &= \frac{1}{m_p}\sum_{v \notin \Sset} p_v w_v .
\end{aligned}
\label{eq:centroids}
\end{equation}
The tail displacements $\Delta^q = c_S^q - c_T^q$ and
$\Delta^p = c_S^p - c_T^p$ measure how far each distribution's tail sits from
its own head.

Throughout, $\Sset$ is fixed at a position and does not depend on $h$. This
holds whenever it is chosen from the teacher, as in any cache written once.
Student-selected sets, used by some on-policy methods, are outside
\cref{thm:residual,thm:invariance}, except that such a set is piecewise constant
in $h$, so both theorems hold within each region where it does not change. They
are not outside the identifiability
threshold, which concerns the teacher alone. We assume $1 \le k < V$, so
$m_p, m_q \in (0,1)$. For the identifiability threshold only, $U$ has full
column rank $d$ and $\mathbf{1} \notin \mathrm{range}(U)$, with $\mathbf{1}$ the
all-ones vector. The second condition is needed because softmax ignores a
constant shift of every logit. Both hold generically when $d \ll V$, and we
verify them per teacher (\cref{tab:e2}).

\section{Theory}
\label{sec:theory}

Two new results with short proofs, and one standard fact. The proofs of the new
results are in \cref{app:proofs}.

\subsection{The objective cannot see tail mass}

\begin{figure}[t]
\centering
\includegraphics[width=0.55\linewidth]{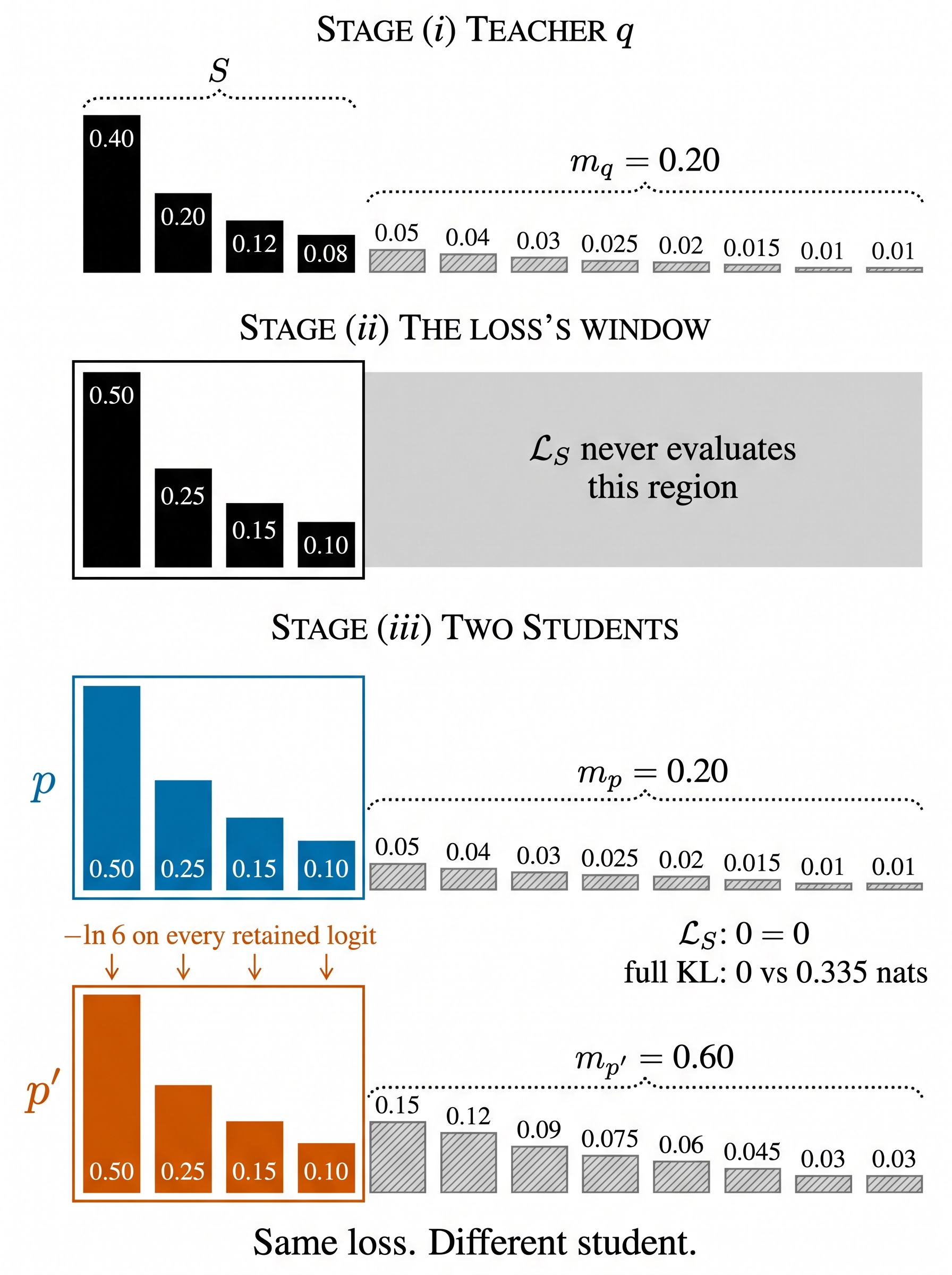}
\caption{Twelve tokens, $k=4$: lowering every retained logit of a student $p$
by $\ln 6$ gives $p'$ with the same renormalized head as $p$ and the teacher
($0.50,0.25,0.15,0.10$), so $\mathcal{L}_\Sset(p)=\mathcal{L}_\Sset(p')=0$. But
$p'$'s tail mass has tripled ($0.20$ to $0.60$) and its full-vocabulary KL to
the teacher is $0.335$ nats against $p$'s $0$.}
\label{fig:same-loss}
\end{figure}

The key observation is that renormalization compares only ratios among retained
tokens. Adding one constant to every retained logit leaves those ratios fixed
but changes the tail mass (\cref{fig:same-loss}; \cref{fig:teaser}a).

\begin{theorem}[The truncated objective does not constrain tail mass]
\label{thm:invariance}
Let $\Sset$ be fixed and independent of $h$, and write
$A = \mathrm{span}\{w_v - w_{v_0} : v \in \Sset\}$, so $\dim A \le k-1$.
Then $\mathcal{L}_\Sset(h)$ depends on $h$ only through $\{a^\top h : a \in A\}$,
so $\mathcal{L}_\Sset(h+\eta) = \mathcal{L}_\Sset(h)$ for every
$\eta \in A^{\perp}$, a subspace of dimension at least $d'-k+1$. Moreover, at
every $h$,
\begin{equation}
\nabla_h m_p \;=\; -\,m_p\,(1-m_p)\,\Delta^p(h) ,
\label{eq:gradmass}
\end{equation}
so the derivative of $m_p$ along $\eta \in A^{\perp}$ vanishes at $h$ if and only
if $\Delta^p(h) \perp \eta$, and $m_p$ is constant on the coset $h + A^{\perp}$
only if $\Delta^p(h') \in A$ throughout it, which is non-generic when $k-1 < d'$.
\end{theorem}

In other words, the loss has no gradient along $A^\perp$, while tail mass
generically does. At $k=64$, $d'=2048$, $A$ spans at most $63$ of $2048$
directions. This is a statement about force, not motion. A parameter step moves
$h$ by $JJ^\top\nabla_h\mathcal{L}_\Sset$ ($J$ the Jacobian of $h$ in the
parameters, times minus the learning rate), which generically leaves $A$. What the theorem rules out is a
restoring force: wherever $h$ moves along $A^\perp$, nothing in the objective
pulls tail mass back. The subspace is per-position, and under shared parameters
training never reaches per-position stationarity; \cref{thm:residual}, which
sums exactly over positions, is the operative result (\cref{sec:discussion}).
\Cref{eq:gradmass} says tail mass moves along $\Delta^p$,
the direction \cref{thm:residual} shows carries the error. Because both theorems concern
$\mathcal{L}_\Sset$ alone, the Mechanism experiment isolates it with
$\alpha_{ce}=0$.

\paragraph{What renormalization does and does not fix.} For a retained set fixed
by the teacher, this is where \citet{sfd2026}'s argument fails: the gradient in the offending direction is
zero, so nothing is redistributed. Their empirical finding stands, since
renormalization does close the cruder loophole of deflating absolute head
probabilities, which is what their ablation isolates. Their $\Sset$ is the
union of student and teacher top-$10$, which depends on $h$, so that case is
open. Ours covers every offline cache and teacher-selected scheme. In our one
student-selected drift comparison, with their union rule, drift was smaller than
under teacher selection ($0.200$ against $0.327$, $p=0.03125$;
\cref{app:torun}).

\subsection{What that blindness costs}

Comparing the two gradients splits the full-vocabulary gradient, at every $h$,
into the part the truncated objective computes and a tail part it never
computes.

\begin{theorem}[Exact gradient decomposition]
\label{thm:residual}
Let $\Sset$ be fixed and independent of $h$. Then for every $h$,
\begin{equation}
\nabla_h \mathcal{L}(h) \;=\; \nabla_h \mathcal{L}_\Sset(h)
\;+\; \big(m_q\, \Delta^q - m_p\, \Delta^p\big) .
\label{eq:residual}
\end{equation}
In particular, at any stationary point $h^\star$ of $\mathcal{L}_\Sset$ the first
term vanishes, so $\nabla_h \mathcal{L}(h^\star) = m_q \Delta^q - m_p \Delta^p$.
That is zero only if $m_p \Delta^p = m_q \Delta^q$; when $m_p = m_q$ it requires
$c_T^p = c_T^q$ exactly.
\end{theorem}

In other words, training reduces what the truncated objective computes and
exerts no force on what it omits, so at a truncated optimum the error is exactly the
tail term (a quantitative lower bound, Corollary~\ref{cor:lowerbound}, is in
\cref{app:proofs}). Removing that term would require reproducing the teacher's
tail centroid with no gradient from the tail (\cref{thm:invariance}).
Stationarity forces the head centroids to agree, not the head distributions.
The two coincide when $\dim A=k-1$, which holds generically for $k-1\le d'$;
\cref{app:results} lists the configurations here where $k-1>d'$ and the blind
subspace is empty.

\subsection{How much of the teacher the retained set shows}

Renormalization cancels the softmax normalizer, so $k$ retained
log-probabilities carry $k-1$ linear constraints on the teacher's $d$ unknowns.
This identifiability threshold is not new
\citep{finlayson2024logits,carlini2024stealing}; we restate it because
\cref{sec:exp} measures it on real teachers.
\label{prop:identifiability}
Let $U$ have full column rank $d$ with $\mathbf{1} \notin \mathrm{range}(U)$, fix
$v_0 \in \Sset$, and set
$D_\Sset = [\,u_v - u_{v_0}\,]_{v \in \Sset \setminus \{v_0\}}
\in \R^{(k-1)\times d}$. Two hidden states give the same renormalized
observation $(\tilde q_v)_{v\in\Sset}$ if and only if they differ by an element
of $\ker D_\Sset$. Hence the states consistent with an observation form an affine
subspace of dimension $d - \mathrm{rank}(D_\Sset) \ge d-k+1$, on which $q$ is not
constant; and $q$ is determined on all of $\Vocab$ exactly when
$\mathrm{rank}(D_\Sset) = d$, which requires $k \ge d+1$, in which case $g$ is
the unique solution of $D_\Sset\, g = \delta_\Sset$, where
$(\delta_\Sset)_v = \log \tilde q_v - \log \tilde q_{v_0} - (b_v - b_{v_0})$.

In other words, below $k=d+1$ a top-$k$ cache cannot pin down the teacher's
full distribution. (Knowing that $g$ leaves a final normalization removes one
dimension, but uniqueness still needs $k=d+1$; \cref{app:proofs}.) This is the
algebra behind output-layer recovery from query access
\citep{finlayson2024logits,carlini2024stealing,stealgeom2026}. Representation
degeneration drives unembedding rows toward a common direction
\citep{gao2019representation,godey2024anisotropy}, which is where $D_\Sset$
becomes ill-conditioned. The threshold is exact arithmetic and says nothing
about conditioning, or about whether the transition is sharp on real
unembeddings. The Threshold experiment measures it.

\section{What to store instead, and what it costs}
\label{sec:method}

\begin{figure}[t]
\centering
\includegraphics[width=0.85\linewidth]{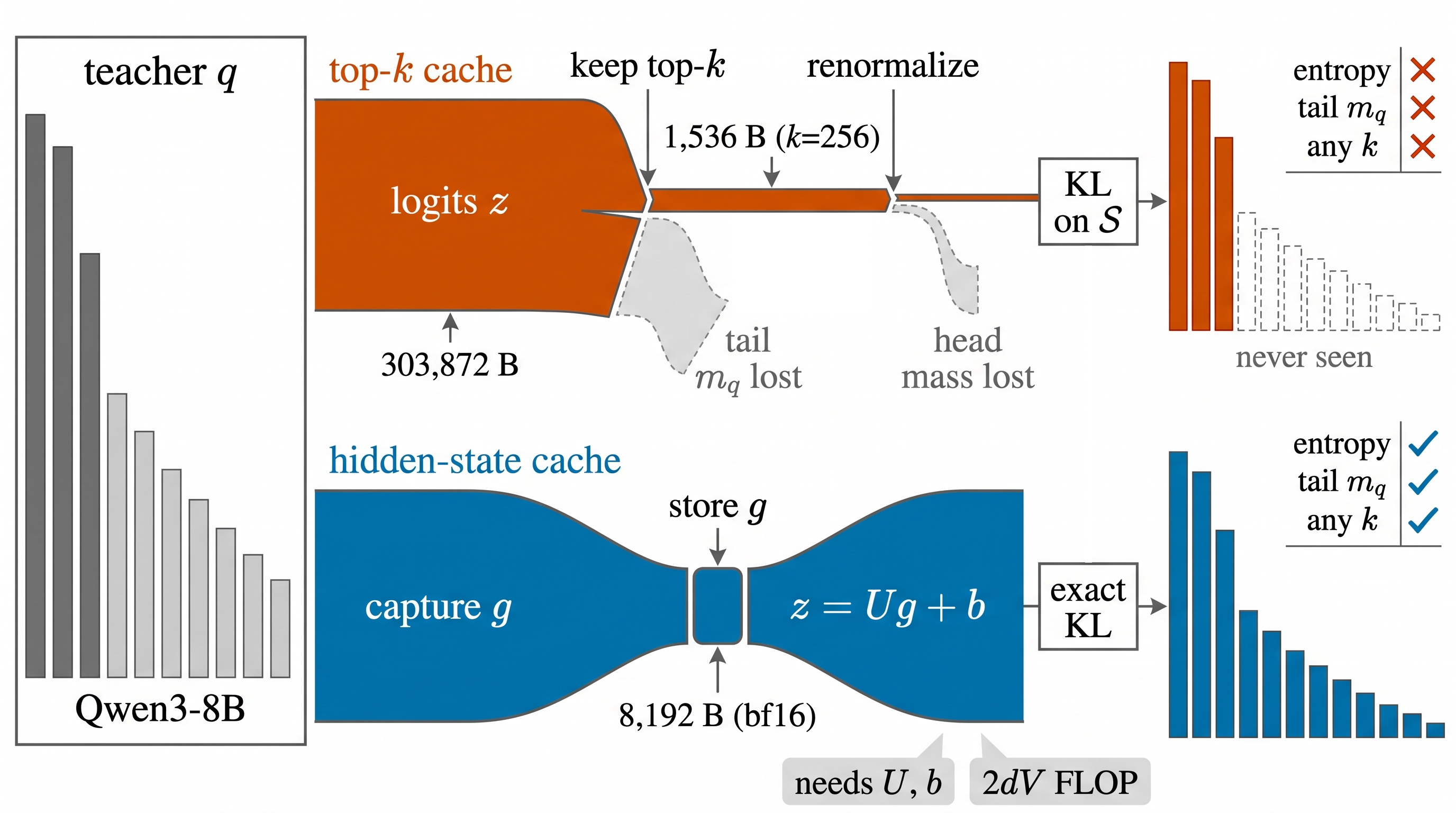}
\caption{A top-$k$ cache crops the teacher's distribution; a hidden-state
cache compresses it. Top: keeping the top-$k$ of the teacher's logits
($1{,}536$ bytes at $k=256$, against $303{,}872$ for all of them) discards the
tail mass $m_q$, and renormalizing discards the head mass $1-m_q$, so the loss
is computed on $\Sset$ alone and the teacher's entropy, $m_q$, and top-$k$ sets
beyond the cached $k$ cannot be recovered. Bottom: caching the final hidden
state $g$ ($8{,}192$ bytes) and reconstructing $z=Ug+b$ gives the exact
full-vocabulary KL and recovers all three, at the cost of access to $U$ and
$b$ and $2dV$ FLOP per token. Qwen3-8B, bf16.}
\label{fig:two-caches}
\end{figure}

If $d+1$ log-probabilities determine the teacher's distribution, the
$d$-dimensional state that generates it is the smallest exact object to cache.
Store the teacher's final hidden state $g$ per supervised position, after the
final normalization, in a low-precision float, and nothing else. At training
time reconstruct exact logits by \cref{eq:logits} inside a fused streaming
divergence kernel \citep{decouplingkl2026,offlinetopk2026}, never materializing
a vocabulary-sized tensor (\cref{fig:two-caches}).

The cache is \emph{exact}, so any divergence against it is the
full-vocabulary divergence. It is \emph{objective-agnostic}: entropy, tail mass
and top-$k$ sets for every $k$ remain recoverable, so one cache serves several
students. It costs the unembedding multiply, $0.6$ to $2.1$ GFLOP per token
across the seven Threshold teachers (\cref{tab:compute}). At bf16 it takes $2d$
bytes per position, against $6k$ for top-$k$ (an int32 index and a bf16 value
per kept token): smaller than top-$1024$ for $d\lesssim3000$, larger than
top-$256$ for every teacher (\cref{tab:storage}). It is the smallest
\emph{exact} cache, not the smallest cache. Reconstruction uses the teacher's
own $U$ at its precision and applies any logit soft-cap afterwards (full
procedure and architecture cases in \cref{app:protocol}).

\section{Experiments}
\label{sec:exp}

Each claim of \cref{sec:theory,sec:method} is tested by at least one
experiment. Three
carry the argument: the blind-subspace probe and Mechanism test
\cref{thm:invariance}, and Threshold tests the identifiability threshold. The
rest support or bound them; two registered predictions are falsified. Predictions marked
\emph{registered} were fixed before the experiment was launched; we report each
as confirmed, partial, falsified or not assessable, and do not reweight after
the fact. \Cref{thm:residual} itself is an identity, verified numerically in
\cref{app:numerics}; the residual-plateau test probes a consequence of it.

\subsection{Scope and protocol}
\label{sec:scope}

Seven teachers, with hidden dimensions $2048$--$8192$ and vocabularies
$128$K--$262$K, are used for Threshold and Systems: Qwen3 1.7B/8B/32B, Llama 3.1
8B/70B, Gemma 3 4B/12B (Gemma 3 adds logit soft-capping, \cref{app:protocol}).
The matched-storage comparison adds Qwen3-4B, Llama-3.2-3B and SmolLM2-1.7B
($V=49$K). Students span 0.36B--1.7B. Off-policy, On-policy and Mechanism use
six seeds, the fewest at which an exact two-sided paired permutation test can
reach $p<0.05$ (floor $2/2^n$). We report Cohen's effect size $d_z$ (bootstrap
intervals in \cref{app:results}); seed-to-seed variation is very small here, so $d_z$ is often
large and should be read with the raw difference. Arms within a comparison share one held-out metric and one
learning rate. The study took about $294$ H200 GPU-hours; hardware, software,
data, compute and convergence checks are in \cref{app:protocol}.

\subsection{Core experiments}

\paragraph{Exactness.} Qwen3-0.6B under Qwen3-8B, 200 steps. At every step we
evaluate the loss against the online teacher and against a bf16 cache of the
same hidden states, on the same student: the two losses agreed to zero
absolute difference at every step. Confirmed, as \cref{eq:logits} requires.

\paragraph{Blind-subspace probe.} Tests \cref{thm:invariance} in a single
backward pass. On Qwen3-8B/Qwen3-0.6B at $256$ real positions, the component of
$\nabla_h\mathcal{L}_\Sset$ in $A^\perp$ is at float precision (median ratio
$2$--$3\times10^{-15}$ of its norm) at every $k$, as \cref{thm:invariance}
requires. The empirical point is the other quantity: the fraction of
$\|\nabla_h m_p\|$ lying in $A^\perp$ is $0.32$, $0.17$ and $0.0027$ at
$k=64,256,1024$, and exactly $0$ at $k=V$, where $A^\perp$ is trivial. Unlike
drift, which noise or initialization could also cause, this isolates the
mechanism directly. Confirmed (\cref{app:torun}).

\paragraph{Threshold.} Registered: error falls to zero at $k=d+1$ and not
before, for every selection rule with full-rank $D_\Sset$. We reconstruct
$\hat q$ from $U\hat g+b$, where $\hat g$ is the least-squares (minimum-norm)
solution of $D_\Sset\,g=\delta_\Sset$, after inverting the cap for soft-capped
teachers. We sweep $k$ from $d/8$ to $4d$; for the soft-capped Gemma teachers
no closed-form solve exists below $d+1$, so their sweep starts there. The median
$\KL(q\|\hat q)$ over $64$ positions is below $10^{-6}$ from $k=d+1$ on all seven
teachers, and on the three Qwen3 teachers it first crosses exactly at $d+1$
(\cref{tab:e2}). Confirmed, with two caveats. For Gemma, ``not before'' is
untested. On Llama-3.1-8B and Llama-3.1-70B the crossing falls one step early, because at $k=d$ the median
$\KL$ is already $1.4\times10^{-7}$ and $3.5\times10^{-8}$. Rank forbids exact
recovery at $k=d$; the early crossing is an artifact of reading one index off
a very steep curve (full curves in \cref{tab:e2-curve}). Every operating point
we found in the literature sits $14$ to $500$ times below its own teacher's
threshold (\cref{tab:cachesizes}).

\begin{table}[t]
\centering
\caption{Threshold: smallest $k$ at which the median $\KL(q\|\hat q)$ over $64$
positions falls below $10^{-6}$. Theory predicts $k=d+1$. Reconstruction is
deterministic. All seven unembeddings have full column rank. Green: collapse at
$k=d+1$; for the soft-capped Gemma teachers smaller $k$ cannot be solved in
closed form and was not measured.}
\label{tab:e2}
\begin{threeparttable}
\small
\setlength{\tabcolsep}{4pt}
\begin{tabular}{@{}l S[table-format=4.0] r S[table-format=1.3] c@{}}
\toprule
Teacher & {$d$} & {Collapse $k$} & {$\mathbf{1}$-res.\tnote{a}} & {$k - (d{+}1)$} \\
\midrule
Qwen3-1.7B    & 2048 & \win{2049} & 0.071 & \win{$0$} \\
Gemma-3-4B    & 2560 & \win{2561} & 0.084 & \win{$0$} \\
Gemma-3-12B   & 3840 & \win{3841} & 0.130 & \win{$0$} \\
Llama-3.1-8B  & 4096 & 4096 & 0.104 & $-1$\tnote{b} \\
Qwen3-8B      & 4096 & \win{4097} & 0.067 & \win{$0$} \\
Qwen3-32B     & 5120 & \win{5121} & 0.062 & \win{$0$} \\
Llama-3.1-70B & 8192 & 8192 & 0.071 & $-1$\tnote{b} \\
\bottomrule
\end{tabular}
\begin{tablenotes}\footnotesize
\item[a] $\min_x\|\mathbf{1}-Ux\|_2/\|\mathbf{1}\|_2$ on $50{,}000$ sampled rows.
\item[b] Tolerance artifact; full curves in \cref{tab:e2-curve}.
\end{tablenotes}
\end{threeparttable}
\end{table}

\begin{table}[t]
\centering
\caption{Operating points of top-$k$ distillation in the literature against
the identifiability threshold $d+1$ of the teacher used (teachers listed in
\cref{app:relatedwork}). Every one sits below it.}
\label{tab:cachesizes}
\small
\setlength{\tabcolsep}{4pt}
\begin{tabular}{@{}l r r r@{}}
\toprule
Study & $k$ & Teacher $d$ & $(d+1)/k$ \\
\midrule
\citet{sfd2026}            & $10$--$20$ & $3584$--$5120$ & $179$--$512$ \\
\citet{rethinkopd2026}     & $16$--$64$ & $2560$--$3584$ & $40$--$224$ \\
\citet{toolcall2026}       & $32$       & $4096$         & $128$ \\
\citet{predistill2024}     & $100$      & $4096$         & $41$ \\
\citet{offlinetopk2026}    & $100$      & $4096$         & $41$ \\
\citet{distillscaling2025} & $128$      & $1792$--$4352$ & $14$--$34$ \\
\bottomrule
\end{tabular}
\end{table}

\paragraph{Mechanism.} The test of \cref{thm:invariance} in the
teacher-selected setting. Registered: at $\alpha_{ce}=0$, top-$k$ students show drift
that tracks initialization and does not decrease with training, while
full-vocabulary students converge to the teacher. Llama-3.2-1B from
Llama-3.1-8B, $k=256$, six seeds: final student tail mass was $0.0784$
(top-$k$) against $0.0471$ (full vocabulary), with the teacher at $0.0216$
($p=0.03125$, $d_z=2031$; \cref{fig:teaser}b). The top-$k$ arm moved away from
the teacher and was still moving at step $500$; the control had flattened
(\cref{app:protocol}). The drifting mass sits on short, very common sub-word
fragments (``re'', ``pre'', ``un''), rarely the single most likely token and so
exactly the tokens that fall just outside a top-$256$ cutoff
(\cref{app:results}). Partial: sign and separation match, but ``tracks
initialization'' and ``converges'' are too strong literally. With the
cross-entropy term restored ($\alpha_{ce}=1$, otherwise identical), the
separation shrinks about fourfold but persists: $0.0542$ against $0.0468$
($p=0.03125$, $d_z=497$), and the top-$k$ arm still ends further from the
teacher than it started (\cref{app:results}).

\paragraph{Drift versus $k$.} If the drift comes from the blind subspace, it
should shrink as $k$ grows. Sweeping $k\in\{16,64,256,1024,4096\}$ at
$\alpha_{ce}=0$, six seeds, final drift falls monotonically
($0.279\to0.143\to0.057\to0.018\to0.0046$); the full-vocabulary arm's,
measured at $k=256$, is $0.0255$. Part of the fall is mechanical, since
the teacher's own tail mass shrinks as $k$ grows. Confirmed.

\paragraph{Residual plateau.} Registered: once $\mathcal{L}_\Sset$ is trained to
convergence, $\|\nabla\mathcal{L}_\Sset\|$ goes to zero while the
full-vocabulary residual $\|\nabla\mathcal{L}\|$ plateaus above it. Neither
holds. Over $5000$ steps (six seeds, standard deviation under $10^{-3}$),
$\|\nabla\mathcal{L}_\Sset\|$ falls from $0.162$ to about $0.134$ within $250$
steps and stays there, and $\|\nabla\mathcal{L}\|$ keeps rising
($0.152\to0.173$). Falsified. Evaluated along the same run, outside its
stationary-point hypothesis, Corollary~\ref{cor:lowerbound}'s lower bound stays below
the residual and rises too ($0.057\to0.079$; $r=0.79$ with the residual over
all checkpoints, $0.9995$ after step $250$). This is consistent with a residual
bounded away from zero rather than a flat one (\cref{app:torun}).

\paragraph{Off-policy.} Registered: caching matches the online arm within
noise, lies on the quality-versus-storage frontier at every budget, and beats
feature matching on held-out perplexity at equal storage. Llama-3.2-1B from
Llama-3.1-8B, six seeds, $8192$ bytes: cached against online perplexity was
$11.656$ against $11.653$ ($p=0.0625$, $d_z=1.28$). The 90\% $t$ interval of the
gap, $[0.0012,0.0055]$, excludes zero, so ``within noise'' is too strong. An
exact equivalence test at $\pm0.01$ perplexity (a margin chosen after the fact,
$2\%$ of the gap to top-$256$) gives $p=0.016$, the one-sided six-seed floor ($1/64$). Both arms receive identical
targets (Exactness), so the gap comes from training itself, where the student's
attention backward is non-deterministic on our hardware; we have not isolated it
further. Cached beat top-$256$, $11.656$ against $12.150$ ($p=0.03125$,
$d_z=-3.35$).

\begin{table}[t]
\centering
\caption{Hidden-state cache against top-$1024$ ($6144$ bytes): held-out
perplexity, six seeds. Every cache-versus-top-$1024$ and
cache-versus-feature-matching difference has $p=0.03125$; $d_z$ is for cache
minus top-$1024$ (negative favors the cache). Green: the better arm in each
row.}
\label{tab:matched}
\small
\setlength{\tabcolsep}{2.5pt}
\begin{tabular}{@{}lrrrrr@{}}
\toprule
Teacher$\to$student & bytes & cache & top-$1024$ & $d_z$ & feature \\
\midrule
Qwen3-1.7B$\to$0.6B   & 4096 & 18.615 & \win{18.336} & $40$    & $83{,}248$ \\
Qwen3-4B$\to$1.7B     & 5120 & 13.885 & \win{13.747} & $25.2$  & $117{,}018$ \\
Gemma-3-4B$\to$1B     & 5120 & \win{14.275} & 15.417 & $-23.6$ & $244{,}352$ \\
Llama-3.2-3B$\to$1B   & 6144 & \win{11.680} & 11.718 & $-6.8$  & $55{,}517$ \\
SmolLM2-1.7B$\to$360M & 4096 & \win{12.558} & 12.569 & $-2.4$  & $684.7$ \\
\bottomrule
\end{tabular}
\end{table}

\paragraph{Off-policy at matched storage.} \Cref{tab:matched} compares the cache
with top-$1024$ at equal or smaller storage on five teachers. The cache wins on
three: widely on Gemma-3-4B, narrowly on Llama-3.2-3B and SmolLM2-1.7B (whose
student, with $d'=960<k-1$, leaves top-$1024$ no blind subspace). It loses on
both Qwen3 teachers. The cache itself is not at fault there. An online
full-vocabulary arm matches it ($18.618$ against $18.615$, $p=0.34$; $13.881$
against $13.885$, $p=0.31$) and loses to top-$1024$ by the same margin, so on
Qwen3 full-vocabulary supervision trains worse than top-$1024$ in this budget.
The median probability outside the top $1024$ is $0.0004$--$0.0006$ for Qwen3,
$0.002$ for SmolLM2 and Llama-3.2, and $0.012$ for Gemma-3-4B. It ranks the
teachers almost as the cache's advantage does (Spearman $0.9$,
\cref{tab:tailmass}). Five teachers is a small sample, and unembedding geometry
also looked predictive at three (\cref{app:results}), so this is a lead, not a
cause. Feature matching fails on every teacher despite a decreasing training
loss. Partial overall: equivalent to online within $0.01$ perplexity, on the
frontier for three of five teachers, and better than feature matching on all
five.

\paragraph{On-policy.} Registered: the gap between truncated and
full-vocabulary supervision persists on-policy and is larger on high-entropy
rollouts. Under GKD \citep{agarwal2024gkd} with forward KL, Qwen3-8B to Qwen3-0.6B,
teacher-selected retained sets, six seeds; positions are split at the batch
median of the teacher's entropy. Mean held-out KL was $0.4513$
(top-$100$) against $0.4182$ (full vocabulary) on low-entropy positions, and
$4.0296$ against $0.5945$ on high-entropy ones (both $p=0.03125$; $d_z=12.7$,
$4.0$). The high-entropy gap is $104\times$ the low-entropy one. Confirmed.

\subsection{Supporting experiments}

\paragraph{Systems, precision, and capacity.} None has a registered
prediction; details in \cref{tab:storage} and \cref{app:results}. Across the
seven Threshold teachers the hidden-state cache is $16$--$102\times$ smaller
than the full logit vector at bf16 ($31$--$205\times$ at int8). Even the
smallest hidden-state cache (int8, $d+4$ bytes) is $5.3$--$21.3\times$ larger
than a top-$64$ cache ($384$ bytes). On a real teacher, int8 reconstruction
error is nearly seven orders of magnitude worse than on a synthetic
teacher ($2.0\times10^{-4}$ against $2.6\times10^{-11}$), so precision cannot
be validated synthetically. Choosing the teacher-student pair matters more than
choosing $k$: at $k=64$, held-out KL is $0.430$, $0.488$ and $0.825$ for
Qwen3-8B$\to$1.7B, Qwen3-1.7B$\to$0.6B and Qwen3-8B$\to$0.6B (a factor of
$1.92$), while raising $k$ from $64$ to $1024$ moves the last only from $0.825$
to $0.731$ (a factor of $1.13$), in these off-policy runs.

\paragraph{Disagreement, falsified.} Registered: tail reliability (how
consistently the teacher ranks its tail tokens across independent text samples,
measured as a rank correlation; \cref{app:protocol}) predicts the sign of the
truncation effect and the ordering of the best $k$, $k^\star$, across three
reproduced settings (task fine-tuning as in \citet{bild2024}, pre-training
distillation, and tool use; \cref{app:protocol}): the less reliable the tail, the smaller $k^\star$. The measured values ($0.455$ bild-style, $0.395$ pretrain-style, $0.181$
toolcall-style) were fixed before the sweep. Held-out $\KL$ fell monotonically with $k$ in all three, and
$k^\star$ was $1024$, the largest value swept: falsified, on a sweep too narrow
to show an interior optimum. Rerun at $k\in\{1024,2048,4096\}$ with six seeds, $\KL$
keeps falling in every setting, each doubling gaining about half as much as the
last (\cref{app:results}). Every step is significant ($p=0.03125$, all six seeds
agreeing), and the gains past $k=1024$ are at most $0.85\%$ of $\KL$. The
hypothesis stays falsified. In no setting, including the stand-in for the one
where truncation was reported to help \citep{bild2024}, was any $k$ up to
$4096$ free on held-out full-vocabulary $\KL$ (not BiLD's task metric),
consistent with \cref{thm:residual} when the tail carries signal.

\section{Discussion and limitations}
\label{sec:discussion}

\paragraph{What survives aggregation over positions.} The results are stated
at one position while training shares parameters across many.
\Cref{thm:residual} aggregates exactly: summed over positions, the
full-vocabulary gradient is the truncated gradient plus per-position tail terms
the objective never evaluates, and this carries the conclusion.
\Cref{thm:invariance} does not. The per-position subspaces differ, and their
intersection $(\sum_i A_i)^{\perp}$ is trivial after roughly $d'/(k-1)$
positions, about $33$ at $k=64$, $d'=2048$. On random unembeddings ($d'=512$,
$k=64$) its dimension falls $449\to386\to260\to9\to0$ at $1,2,4,8,9$
positions; in a separate check, summing \cref{eq:residual} over $40$ positions
reproduces the aggregate gradient within $10^{-14}$ (\cref{app:results}). What survives
is that the objective never evaluates tail mass, exactly the point at issue
with \citet{sfd2026}. The identifiability threshold bounds what one observation
determines of the teacher's \emph{state}, not what a student learns of its
\emph{function} across positions.

\paragraph{Does the cross-entropy term make this moot?} Standard offline
recipes keep $\alpha_{ce}>0$. \Cref{thm:residual} concerns $\mathcal{L}_\Sset$
alone, whatever else is in the loss. Hence whatever holds tail mass in place
there comes from the cross-entropy term, not from distillation. In Mechanism it
narrows the gap about fourfold without closing it (\cref{sec:exp}). In a recipe
that drops or heavily downweights that term, nothing in the objective holds
tail mass.
The On-policy experiment trains on the forward KL alone, as \citet{sfd2026} do.
Its gaps (truncation adds $8$ percent to the full-vocabulary arm's KL on
low-entropy positions and $578$ percent on high-entropy ones) are measured in
exactly that regime.

\paragraph{We do not explain the published disagreement.} Our results describe
what truncated supervision cannot constrain, not which settings it hurts.
\Citet{geomlimits2026} argue that a student's representational width bounds
the features it can encode, whatever the objective, consistent with our
Capacity result. Holding the pair fixed and changing only the corpus moves
the level of $\KL$ but not whether an optimum exists (\cref{app:results}).

\paragraph{We claim no method, and the remedy has limits.} Hidden-state caching
with prediction-head reconstruction is deployed practice
\citep{deepseekv4,kdflow2026}. It needs $U$, $b$ and the ability to export $g$,
which excludes a black-box API; it is larger than a top-$64$ cache
(\cref{tab:storage}); and it helps on storage only off-policy
\citep{opdsurvey2026}.

\paragraph{Limitations.} The practical payoff of caching is teacher-dependent
(\cref{tab:matched}), and five teachers cannot establish why. The theorems
cover student-selected sets only where the set does not change; one comparison
(\cref{sec:theory}) does not close that gap. None of this touches
\cref{thm:invariance,thm:residual} (proved) or the cited identifiability
threshold, all verified numerically (\cref{app:numerics}). Cross-tokenizer alignment \citep{dskd2025}, a frozen
reference policy's KL in reinforcement learning, multi-teacher caching, long
context, and the supervised-position fraction remain untested.

\section{Conclusion}
\label{sec:conclusion}

Renormalization has been assumed to make the retained view a faithful target.
It does not. The renormalized objective sees only head log-odds, and its
gradient misses exactly one tail term. Hence a student can match its teacher on
everything it is shown and still be wrong by an amount its objective will not
reduce. Off-policy, the effect is modest: changing the teacher-student pair moved
held-out divergence by a factor of $1.9$, changing $k$ by $1.1$: in these runs
the pair mattered more than $k$. On-policy, on high-entropy positions, truncation
costs far more.
Renormalization is a normalization, not a safeguard.

\begin{kilrepro}
All theoretical claims are stated with explicit assumptions in
\cref{sec:theory}. The two theorems and the corollary are proved in
\cref{app:proofs}; the identifiability threshold is a known result, cited
rather than reproved. All are re-derived numerically in \cref{app:numerics}. The code is available at \url{https://github.com/KurbanIntelligenceLab/top-k-distillation}. It contains the full distillation, caching and
reconstruction code, the job script behind every reported run, the per-run JSON
results each reported number is drawn from, and a README mapping each
experiment to its scripts and results. A script in the repository re-derives from
those files every reported quantity that is checkable without rerunning a model,
and prints a pass or fail for each; it needs only NumPy and runs on a CPU in
seconds. Hardware, software versions, datasets and compute are listed in
\cref{app:protocol}.
\end{kilrepro}




\section*{Ethics statement}
This work uses the structural results underlying model-extraction attacks in a
cooperative setting, where the teacher's weights are already available to the
party performing distillation; it assumes white-box access and does not lower
the cost of extraction from query access. Standard distillation concerns apply,
including propagation of a teacher's biases and errors. Exact supervision
propagates the teacher's tail behavior more faithfully, which helps calibration
and hurts if the tail encodes harmful content.

\subsection*{AI use statement}
An AI assistant (Claude, Anthropic) was used under author direction to help
formalize and draft proofs, implement and debug the experimental code, extract
numbers from result files, and edit text, figures and references. The
theoretical results, the experiments and registered predictions, and the
interpretation of every result are the authors'. AI-drafted proofs were
re-derived by hand by at least two authors, and the authors take full
responsibility for all content.

\bibliography{references}

\appendix

\section{Proofs}
\label{app:proofs}

The identifiability threshold is the rank argument behind output-layer recovery
\citep{finlayson2024logits,carlini2024stealing}, and we do not reprove it. We add
one observation. Whoever holds the teacher also knows that $g$ leaves a final
normalization: for an RMSNorm with gain $\gamma$, $g$ lies on the ellipsoid
$\sum_i (g_i/\gamma_i)^2 = d$ (ignoring the norm's small $\epsilon$). At $k=d$
the line of states consistent with the observation meets this ellipsoid in at
most two points, which generically give different $q$, so a unique answer still
needs $k=d+1$. Below we prove the two new results and the corollary.

\paragraph{Proof of \cref{thm:invariance}.} For $v\in\Sset$,
$\tilde p_v = e^{w_v^\top h}/\sum_{u\in\Sset}e^{w_u^\top h}$ depends on $h$ only
through $(w_v-w_{v_0})^\top h$. So $\mathcal{L}_\Sset$ depends on $h$ only through
$\{a^\top h : a\in A\}$ and is constant along $A^\perp$, with
$\dim A^\perp \ge d'-k+1$ because $\dim A\le k-1$.
For \cref{eq:gradmass}, let $\bar w=\sum_u p_uw_u$. Since
$\nabla_h p_v = p_v(w_v-\bar w)$ and $\sum_{v\in\Sset}p_v=1-m_p$,
\[
\nabla_h m_p = -\sum_{v\in\Sset}p_v(w_v-\bar w) = -(1-m_p)(c_S^p-\bar w),
\]
and $\bar w=(1-m_p)c_S^p+m_pc_T^p$ gives $c_S^p-\bar w = m_p\Delta^p$.
The derivative of $m_p$ along $\eta\in A^\perp$ is therefore
$-m_p(1-m_p)\langle\Delta^p(h),\eta\rangle$, which vanishes if and only if
$\Delta^p(h)\perp\eta$. Constancy of $m_p$ on the coset $h+A^\perp$ needs
$\Delta^p(h')\in A$ at every $h'$ in it. Since $c_S^p(h')$ lies in
$w_{v_0}+A$, this holds exactly when $c_T^p(h')\in w_{v_0}+A$, a proper affine
subspace when $k-1<d'$. We call this non-generic without claiming a measure-zero
statement, since the law of $c_T^p$ is not under our control. \qed

\paragraph{Proof of \cref{thm:residual}.} Using $\sum_v q_v=1$,
$\nabla_h \KL(q\|p) = \sum_v(p_v-q_v)w_v$, and the same computation on $\Sset$
gives $\nabla_h \mathcal{L}_\Sset = c_S^p - c_S^q$. On $\Sset$,
$p_v=(1-m_p)\tilde p_v$ and $q_v=(1-m_q)\tilde q_v$; off $\Sset$ the sums are
$m_pc_T^p$ and $m_qc_T^q$. With $c_T=c_S-\Delta$ for each distribution,
\[
\nabla_h \mathcal{L}
= (1-m_p)c_S^p-(1-m_q)c_S^q + m_pc_T^p - m_qc_T^q
= c_S^p - c_S^q + m_q\Delta^q - m_p\Delta^p ,
\]
which is \cref{eq:residual}. At a stationary point the first term vanishes. If
$m_p=m_q>0$, the remainder vanishes if and only if $\Delta^p=\Delta^q$, that is,
$c_T^p=c_T^q$, since the head centroids then coincide. \qed

\paragraph{A lower bound on the residual.} Under a separation condition on the
tail displacements, the residual at a stationary point is bounded below.

\begin{corollary}[Lower bound]
\label{cor:lowerbound}
Let $h^\star$ be a stationary point of $\mathcal{L}_\Sset$, so that
$c_S^p = c_S^q =: c_S$ and $\nabla_h\mathcal{L}(h^\star) = m_q\Delta^q -
m_p\Delta^p$ by \cref{thm:residual}. Suppose $c_S \neq 0$ and, writing
$\hat c_S = c_S/\|c_S\|$, that there are $0 < \gamma \le \Gamma$ with
$\langle \Delta^q, \hat c_S\rangle,\ \langle \Delta^p, \hat c_S\rangle
\in [\gamma, \Gamma]$. Then
\begin{equation}
\big\| \nabla_h \mathcal{L}(h^\star) \big\|_2 \;\ge\; m_p\,\gamma - m_q\,\Gamma ,
\label{eq:lowerbound}
\end{equation}
which is strictly positive whenever $m_p / m_q > \Gamma/\gamma$.
\end{corollary}

\paragraph{Proof of Corollary~\ref{cor:lowerbound}.} Since $\|\hat c_S\|=1$,
$\|\nabla_h\mathcal{L}(h^\star)\|_2 \ge
\langle\nabla_h\mathcal{L}(h^\star), -\hat c_S\rangle
= m_p\langle\Delta^p,\hat c_S\rangle - m_q\langle\Delta^q,\hat c_S\rangle
\ge m_p\gamma - m_q\Gamma$. \qed

\section{Numerical verification of the theory}
\label{app:numerics}

A dedicated script re-derives all three results numerically
from scratch, NumPy only, in about a minute on CPU;
another repeats the gradient identities
symbolically in SymPy and probes the boundary cases $k=1$ and $k=V$. Both compute quantities
independently of the paper's reported numbers and assert nothing about
outcomes; they print what they measure.

\paragraph{The identifiability threshold.} For a random unembedding with
$V=4000$, $d=64$, reconstructing $g$ from $k$ observed log-probabilities under
top-$k$ selection: $\mathrm{rank}(D_\Sset)$ equals $\min(k-1,d)$ and the fiber
dimension equals $d-r$ at every $k$, as stated above. $\KL(q\|\hat q)$
is $3.4\times10^{-2}$ at $k=d$ and $4.2\times10^{-17}$ at $k=d+1$, and the
maximum log-probability error falls from $10.2$ to $1.5\times10^{-13}$ across the
same unit increase: the transition sits exactly at the threshold and not one
index earlier. Random selection behaves identically ($1.0\times10^{-1}$ at $k=d$,
$9.3\times10^{-17}$ at $k=d+1$). Top-$k$ selection does not collapse early here, unlike
the two real Llama teachers (\cref{tab:e2-curve}).

\paragraph{\Cref{thm:residual}.} The decomposition is verified symbolically as
an identity in $h$ and numerically at random $h$ far from any optimum
($V=800$, $d'=9$, $k=4$; residual $\le 8\times10^{-15}$ while
$\|\nabla\mathcal{L}_\Sset\| \approx 3$), so it is not an artifact of
evaluating near stationarity. Evaluating the stationary-point form at an
approximate stationary point leaves a discrepancy of exactly
$\nabla_h\mathcal{L}_\Sset$, by \cref{eq:residual} itself; this is why the
measured mismatch tracks $\|\nabla\mathcal{L}_\Sset\|$ one for one over seven
decades ($9.97\times10^{-5} \to 9.76\times10^{-5}$;
$9.97\times10^{-12} \to 9.73\times10^{-12}$) rather than approximately. For
$V=3000$, $d'=48$, $k=20$ at $\|\nabla\mathcal{L}_\Sset\| < 10^{-11}$,
Corollary~\ref{cor:lowerbound} was valid in every trial and recovered $29$--$37\%$ of the
true residual norm, so the bound is non-vacuous.

\paragraph{\Cref{thm:invariance}.} For $V=2000$, $d'=32$, $k=12$:
$\dim A^\perp = 21 = d'-k+1$ exactly; $\mathcal{L}_\Sset$ is invariant along
$A^\perp$ to machine precision ($|\Delta\mathcal{L}_\Sset| \le 10^{-15}$); and
\cref{eq:gradmass} matches central finite differences to a relative error of
$4$--$5\times10^{-8}$ across four seeds, while $m_p$ itself changes by
$10^{-4}$ along the same direction, which is the content of the theorem.

These checks confirm the algebra numerically; at least two authors have
additionally re-derived each proof in \cref{app:proofs} line by line by
hand, independent of this numerical verification.

\section{Extended related work}
\label{app:relatedwork}

\Cref{sec:related} states the positioning; this appendix adds only what does not
fit there.

\paragraph{Per-study detail on truncated supervision.} \Citet{duet2026} report
degradation only once $k$ is raised far enough, the opposite ordering to
\citet{sparselogit2025}. \Citet{distillscaling2025} give the most controlled
measurement we found: at $k=128$ against $k=32768$, validation loss rises by
$0.11$ nats, and top-$p$ at $p=0.9$ costs $0.13$ nats against $p=1.0$.
\Citet{toolcall2026} report that a teacher's top-$32$ retains $99.99$ percent of
probability mass while containing the decision-critical token on only $0.4$
percent of audited prompts, which is the sharpest available demonstration that
retained mass is a poor proxy for retained supervision. \Citet{offlinetopk2026}
report that an offline top-$k$ cache matches online distillation at
near-identical training loss. This is the closest published challenge to our
motivation. We do not contradict it: matching online top-$k$
training loss is a different statement from matching full-vocabulary quality.

\paragraph{Reported cache sizes against the identifiability threshold.}
\Cref{tab:cachesizes} lists the operating point of each study cited for its
choice of $k$, with its teacher's hidden dimension $d$ (from each model's public
configuration). The teachers are: Skywork-OR1-Math-7B, DeepCoder-14B and
R1-Distill-Qwen-32B \citep{sfd2026}, whose $k$ is the union of student and
teacher top-$10$; R1-Distill-Qwen-7B, Qwen3-4B and Skywork-OR1-Math-7B, at
$k=16$ with a sweep to $64$ \citep{rethinkopd2026}; 9B teachers trained from
Qwen3.5-9B \citep{toolcall2026}; GLM-4-9B, after top-$p$ truncation at $0.95$
\citep{predistill2024}; Llama-3.1-8B-Instruct \citep{offlinetopk2026}; and the
authors' own 546M, 1.82B and 7.75B models \citep{distillscaling2025}. Sweeps can go further:
\citet{distillscaling2025} sweep $k$ up to $4096$, above $d+1$ for two of their
three teachers, but report the cost of truncation at $k=128$.

\paragraph{Feature distillation and statistical views.} An older line transfers
intermediate representations rather than output distributions
\citep{romero2015fitnets,wang2020minilm}. Feature matching constrains the
student's representation to resemble the teacher's, and needs a dimension-matching
projection. Reconstruction, by contrast, recovers the output distribution exactly and
leaves the representation unconstrained. Feature matching is the arm that isolates exact
reconstruction from mere hidden-state availability, and it is
not competitive at all (\cref{sec:exp}). \Citet{menon2021statistical} analyze distillation as variance
reduction from using teacher probabilities in place of sampled labels, with a
bias-variance account in which an imperfect teacher can beat a more accurate but
higher-variance one. That is the tradition our falsified hypothesis belonged to,
with truncation as the controlled variable instead of teacher capacity. We could
not state it as a theorem: independent perturbations of the teacher's
log-probabilities do not lie in the image of $U$, so they violate
\cref{eq:logits}, and we found no rank-consistent noise model strong enough to
give an interior optimum in $k$. \Citet{infoview2026} obtain a bias, variance
and rank-bottleneck decomposition in a linear Gaussian case, but for the
student's generalization gap, not for truncated supervision.

\section{Extended experimental protocol}
\label{app:protocol}

\paragraph{Scope, compute, software, and data.} The theory constrains any
supervision transmitting $k<d+1$ teacher log-probabilities from a fixed
retained set, however the sequences arose; an offline cache additionally
offers a storage advantage only off-policy, since a fully on-policy teacher is
in the loop by construction. Hardware: every experiment ran on a single NVIDIA
H200 GPU per job (16 CPU cores, 100GB RAM), except the Llama-3.1-70B Threshold
run, which needed two H200s sharded via \texttt{device\_map="auto"} (its bf16
weights alone are within 1GB of a single H200's capacity). We estimate the
total compute cost of every experiment reported in this paper at about $294$
GPU-hours: $243$ from the cluster's own job-scheduler accounting (smoke tests
included), plus wall-clock-based estimates of $23$ GPU-hours for the 70B
Threshold run and its two smoke tests and $27$ for the $\alpha_{ce}=1$
Mechanism rerun and its smoke test, and $1$ for the two Qwen3 follow-ups (online
arm and tail mass) and their smoke test. Per-experiment GPU-hours: Exactness $0.2$; Threshold, including the
drift-versus-$k$, student-selection, residual-plateau design tests, an
$\alpha_{ce}$ sweep invalidated by an implementation error and not reported, and the 70B run $154.3$; Off-policy
$3.6$; On-policy $29.6$; Mechanism, including the token-absorption pass
$71.5$; Systems $0.03$; Disagreement $1.1$; Precision $0.1$; Capacity $1.4$;
one-time environment setup $0.8$; interactive debugging during development
$3.3$; the $\alpha_{ce}=1$ Mechanism rerun $27$; the two Qwen3 follow-ups $1$. Software: Python 3.11.15, PyTorch 2.13.0+cu130, Transformers 5.15.1,
\texttt{datasets} 5.0.1, NumPy 2.4.6. Datasets: FineWeb-Edu
(\texttt{HuggingFaceFW/fineweb-edu}, ODC-BY), UltraChat-200k
(\texttt{HuggingFaceH4/ultrachat\_200k}, MIT), FLAN's CNN/DailyMail task shard
(\texttt{Muennighoff/flan}, Apache-2.0), and Glaive Function Calling v2
(\texttt{glaiveai/glaive-function-calling-v2}, Apache-2.0). Models: Qwen3 and
SmolLM2 (Apache-2.0), Llama 3.1 and 3.2 (Llama Community License), Gemma 3
(Gemma Terms of Use). Each corpus is drawn
from a single fixed shard, shuffled once by seed, with an exact-length filter
after tokenization, documented in the data-loading code in the repository. Training runs use
batch $4$ at sequence length $512$, except Exactness (sequence length $1024$) and
On-policy (batch $8$, $256$-token prompts, $256$ generated tokens); steps are
$200$ (Exactness), $100$ (Disagreement), $150$ (Capacity), $5000$ (residual
plateau) and $500$ elsewhere. No experiment tunes the token budget per arm.

\paragraph{Protocol checks.} Metric identity across arms: every quality
comparison evaluates all its arms on the same held-out metric over the same
support, computed identically regardless of what each arm trained on. The
On-policy $104\times$ figure in particular is the same exact full-vocabulary
forward-KL, evaluated under no gradient, for both the truncated and the exact
arm; only the training objective differs. Hyperparameters: every comparison
uses a single constant learning rate and a plain AdamW optimizer with no warmup
and no schedule, passed identically to every arm within one script invocation
(Off-policy, including every matched-storage run: $2\times10^{-5}$;
Mechanism, On-policy, Disagreement and Capacity: $10^{-5}$); no comparison tunes a hyperparameter on one arm and reuses it
for another. Convergence: the Mechanism control (full-vocabulary supervision) is
not a still-moving transient. Its per-step rate of tail-mass change falls by
about $33\times$ from the first $50$ of $500$ steps to the last $250$
($-4.0\times10^{-5}$ per step early, $-1.2\times10^{-6}$ per step late) and is
effectively flat from step $50$ onward. The truncated arm's rate falls by a
similar factor over the same window ($3.4\times10^{-4}$ early to
$9.4\times10^{-6}$ late, about $36\times$), but its late rate is about
$8\times$ the control's and of opposite sign: it has not plateaued, and
continues to move away from the teacher through step $500$. The
comparison is therefore between a converged control and a still-drifting
truncated arm, the qualitative signature \cref{thm:invariance} predicts, not an
artifact of neither arm having settled.

\begin{table}[h]
\centering
\caption{Head cost of reconstructing logits from $g$: $2dV$ FLOP per
token.}
\label{tab:compute}
\begin{threeparttable}
\begin{tabular}{l S[table-format=4.0] S[table-format=6.0] S[table-format=1.2]}
\toprule
Teacher & {$d$} & {$V$} & {Head (GFLOP)} \\
\midrule
Qwen3-1.7B    & 2048 & 151936 & 0.62 \\
Gemma-3-4B    & 2560 & 262144 & 1.34 \\
Gemma-3-12B   & 3840 & 262144 & 2.01 \\
Llama-3.1-8B  & 4096 & 128256 & 1.05 \\
Qwen3-8B      & 4096 & 151936 & 1.24 \\
Qwen3-32B     & 5120 & 151936 & 1.56 \\
Llama-3.1-70B & 8192 & 128256 & 2.10 \\
\bottomrule
\end{tabular}
\end{threeparttable}
\end{table}

\paragraph{Off-policy arm list, as scoped.} The eight arms as scoped, to be
compared at matched storage and separately at matched training FLOPs: (i) cross-entropy only;
(ii) forward-KL top-$k$ ($k \in \{16,64,256,1024\}$) and top-$p$, current
practice; (iii) BiLD \citep{bild2024}; (iv) reverse KL and JSD/total variation
\citep{gu2024minillm,wen2023fdivergence}; (v) skew KL \citep{ko2024distillm};
(vi) hidden-state feature matching \citep{romero2015fitnets,wang2020minilm};
(vii) online full-vocabulary supervision, the exact-quality upper reference;
(viii) hidden-state caching. Each isolates a confound: BiLD and top-$p$
represent deployed truncation practice; reverse KL and JSD test whether
\cref{thm:residual}'s residual is specific to forward KL's mode-covering
behavior; skew KL is a strong off-policy-tolerant baseline; feature matching
also uses the teacher's hidden state but regresses against it directly rather
than reconstructing logits, isolating the benefit of exact reconstruction from
mere hidden-state availability. Arms (i), (iii)--(v) and top-$p$ were not run; of (ii), only top-$256$ and
top-$1024$ were run; and no matched-FLOP comparison was run.
Feature matching, run at six seeds on every matched-storage teacher
(\cref{tab:matched}), is not competitive at all (on Qwen3-1.7B, $83{,}248$
perplexity against $18.3$--$18.6$ for cached and top-$1024$, $p=0.03125$)
despite a training loss that visibly decreases, showing that mere access to
the teacher's hidden state, without exact reconstruction, does not by itself
yield a usable language model. This rules out the competing explanation the
arm was designed to test.

\begin{table}[h]
\centering
\caption{Hidden-state cache size per supervised position, and its ratio to the
full logit vector and to a top-$64$ cache. Green: how much smaller than the
full logits the exact cache is.}
\label{tab:storage}
\begin{threeparttable}
\begin{tabular}{l S[table-format=4.0] S[table-format=6.0] S[table-format=5.0] S[table-format=4.0] r S[table-format=2.2]}
\toprule
& & & \multicolumn{2}{c}{Hidden-state cache (B)} & \multicolumn{2}{c}{Ratio ($\times$)} \\
\cmidrule(lr){4-5}\cmidrule(lr){6-7}
Teacher & {$d$} & {$V$} & {bf16} & {int8\tnote{a}} & {vs.\ full} & {vs.\ top-64} \\
\midrule
Qwen3-1.7B    & 2048 & 151936 &  4096 & 2052 & \win{74.2} &  5.34 \\
Gemma-3-4B    & 2560 & 262144 &  5120 & 2564 & \win{102.4} &  6.68 \\
Gemma-3-12B   & 3840 & 262144\tnote{b} &  7680 & 3844 & \win{68.3} & 10.01 \\
Llama-3.1-8B  & 4096 & 128256 &  8192 & 4100 & \win{31.3} & 10.68 \\
Qwen3-8B      & 4096 & 151936 &  8192 & 4100 & \win{37.1} & 10.68 \\
Qwen3-32B     & 5120 & 151936\tnote{b} & 10240 & 5124 & \win{29.7} & 13.34 \\
Llama-3.1-70B & 8192 & 128256 & 16384 & 8196 & \win{15.7} & 21.34 \\
\bottomrule
\end{tabular}

\begin{tablenotes}\footnotesize
\item[a] Symmetric max-abs, one fp32 scale: $d+4$ bytes. ``vs.\ full'': bf16
cache against bf16 logits ($2V$ bytes); ``vs.\ top-64'': int8 cache against
$384$ bytes.
\item[b] From the published config, checked against the checkpoint.
\end{tablenotes}
\end{threeparttable}
\end{table}

\paragraph{Disagreement experiment design.} The three reproduced settings,
task-specific fine-tuning after \citet{bild2024}, pre-training distillation after
\citet{sparselogit2025,predistill2024}, and the tool-use setting of
\citet{toolcall2026}, are the primary sources of the published disagreement.
Reproducing them under one protocol, rather than citing original numbers,
removes confounds from differing model families, data, and evaluation suites.
Each reproduction substitutes an open, downloadable stand-in for the original
setting's proprietary or unavailable data/models, matched by task category
rather than by exact source: task-specific fine-tuning
(\citet{bild2024}'s setting) is approximated by Qwen3-8B/Qwen3-0.6B on FLAN's
instruction/response pairs (\texttt{Muennighoff/flan}); pre-training
distillation (\citet{sparselogit2025,predistill2024}'s setting) by
Llama-3.1-8B/Llama-3.2-1B on general web text
(\texttt{HuggingFaceFW/fineweb-edu}); and the tool-use setting
(\citet{toolcall2026}) by Qwen3-8B/Qwen3-0.6B on tool-calling conversations
(\texttt{glaiveai/glaive-function-calling-v2}). All three run under one
shared protocol: $k \in \{8,64,256,1024\}$, three seeds, $100$ training steps,
batch $4$ at sequence length $512$, tail reliability registered from $32$
predictor batches before the sweep. This is not a claim that the
substituted data or models match the originals' domains exactly (e.g.\ FLAN
is a proxy for whatever proprietary task-tuning corpus \citet{bild2024} used,
not a reproduction of it), only that the truncation-versus-tail-reliability
mechanism can be tested under one controlled protocol across three settings
that differ in the same qualitative way the disagreement in the literature
does (task-specific vs.\ general-purpose vs.\ decision-critical-token
supervision). The falsification (\cref{sec:exp}) is therefore a statement
about this mechanism under this protocol, not a claim that any individual
original result was wrong.

\paragraph{The write and read procedure in full.} An expert should be able to
reproduce this without guessing, so we state both paths explicitly. Symbols are
as in \cref{sec:setup-notation}.

Write (once, per supervised position).
(i) Run the teacher forward and capture $g \in \R^{d}$ at the residual stream
after the final normalization, including any learned gain and any output
scaling the architecture applies before the head, and before the
unembedding matrix multiply. For an RMSNorm head with learned gain $\gamma$, $g$
is the post-gain vector, so that \cref{eq:logits} holds with the model's own $U$.
(ii) Quantize. For bf16, store $g$ directly at $2d$ bytes with no scale. For
int8, use symmetric max-abs quantization, $s = \max_i |g_i| / 127$ and
$\hat g_i = \mathrm{clip}(\mathrm{round}(g_i/s), -127, 127)$, storing the $d$
int8 values and one fp32 $s$, for $d+4$ bytes; dequantize as $s \hat g$.
(iii) Store nothing else: no probabilities, no logits, no index array. The
teacher's $U$ and $b$ are stored once per teacher, not per token.

A dedicated script checks this procedure end to end: the
byte accounting, the quantizer, and the order of operations for soft-capping.

Read (each training step).
(i) Dequantize $g$ to the student's compute dtype.
(ii) Form $z = Ug + b$ in chunks over the vocabulary inside a fused streaming
divergence kernel \citep{decouplingkl2026,offlinetopk2026}, never materializing a
$V$-sized tensor. Use the same $U$ at the same precision the teacher used; if the
teacher was quantized, reconstructing with a higher-precision $U$ makes the logit
error dominated by $U$'s quantization rather than by $g$'s.
(iii) Apply any post-head transformation the teacher applies, in the teacher's
order. For logit soft-capping this means computing $c\tanh(z/c)$ after
reconstruction; for temperature $\tau$, dividing by $\tau$, which commutes with
reconstruction and can equivalently be folded into the divergence.
(iv) Compute the divergence against the student's logits. Because
\cref{eq:logits} is exact, the result is the full-vocabulary divergence, not an
approximation, which is what the Exactness experiment checks
(zero absolute difference at every step).

\paragraph{Architectural cases requiring care, in full.} Final
normalization: $g$ must be captured after the final RMSNorm or LayerNorm and any
learned output scaling. Tied embeddings, temperature scaling, and
mixture-of-experts: no change needed. With tied embeddings, $U$ is the input
embedding table itself; with temperature scaling, the scalar applied to the
logits commutes with reconstruction; and with mixture-of-experts, routing
affects the body rather than the head. Logit soft-capping:
$z\mapsto c\tanh(z/c)$ is monotone and known, so reconstruction applies to the
pre-cap logits and the cap is reapplied. Mixture-of-Softmaxes heads:
these deliberately break the rank constraint
\citep{yang2018breaking,ganea2019breaking}, raising the identifiability
threshold above; caching all mixture components' hidden states
restores exactness. Quantized teachers: reconstruction must use the same
$U$ at the same precision the teacher used, or logit error is controlled by
$U$'s quantization error rather than $g$'s.

\section{Extended results}
\label{app:results}

\begin{table}[h]
\centering
\caption{Median $\KL(q\|\hat q)$ near the transition for the two teachers
marked $-1$ in \cref{tab:e2}: already below $10^{-6}$ at $k=d$ but not zero,
and floating-point noise from $k=d+1$ on (negative entries are rounding).
Green: exact recovery.}
\label{tab:e2-curve}
\begin{tabular}{l r r}
\toprule
$k$ & Llama-3.1-8B ($d=4096$) & Llama-3.1-70B ($d=8192$) \\
\midrule
$0.90d$ & $9.98\times 10^{-5}$   & $2.06\times 10^{-5}$ \\
$d$     & $1.39\times 10^{-7}$   & $3.48\times 10^{-8}$ \\
$d+1$   & \win{$2.72\times 10^{-8}$}   & \win{$2.45\times 10^{-8}$} \\
$d+2$   & \win{$-1.48\times 10^{-8}$}  & \win{$5.75\times 10^{-9}$} \\
$2d$    & \win{$-2.59\times 10^{-10}$} & \win{$-2.90\times 10^{-9}$} \\
\bottomrule
\end{tabular}
\end{table}

\paragraph{Threshold: the early crossing on the two Llama teachers.} At $k=d=4096$ (Llama-3.1-8B) and $k=d=8192$ (Llama-3.1-70B) the
median $\KL$ is $1.4\times10^{-7}$ and $3.5\times10^{-8}$, below the $10^{-6}$
tolerance, while $k=d+1$ gives $2.72\times10^{-8}$ and $2.45\times10^{-8}$,
within an order of magnitude of the $k=d$ value at either scale. The
identifiability threshold forbids \emph{exact} recovery at $k=d$, since
$\mathrm{rank}(D_\Sset)\le d-1$ there, and the measurement does not contradict
this: $\KL$ at $k=d$ is small but not exactly zero, close to the
$10^{-8}$--$10^{-10}$ floating-point floor of the $k\ge d+1$ rows, however far
past $d+1$ they are. The apparent early crossing comes from reporting a single
threshold-crossing index on a curve that is already very steep just below the
true transition, not from an off-by-one in the sweep grid; that it recurs at two
teacher scales strengthens this reading. \Cref{tab:e2-curve} gives
both curves.

\paragraph{Threshold: selection-rule breakdown.} Under random and
leverage-score selection the collapse sits at $d+1$ for every teacher (for
Gemma, measured from $d+1$ only); under
top-$k$ selection it does too, except for the two Llama teachers discussed above
(\cref{tab:e2}). Selection changes how the error falls before the threshold, not
where the threshold is. For the soft-capped Gemma teachers, reconstruction
inverts the cap and searches over the unknown normalizer in one dimension. For
scale, at $k=256$ under top-$k$ selection Qwen3-1.7B still has
$\KL(q\|\hat q)=0.485$, nearly six orders of magnitude above the $10^{-6}$
tolerance.

\paragraph{Configurations with an empty blind subspace.} Stationarity of
$\mathcal{L}_\Sset$ forces the head centroids to agree, $c_S^p=c_S^q$. Since
both heads sum to one, this reads $\sum_{v\in\Sset}(\tilde p_v-\tilde q_v)(w_v-w_{v_0})=0$,
which forces $\tilde p=\tilde q$ exactly when $\dim A=k-1$. That holds
generically when $k-1\le d'$, which covers every configuration here except the
widened Disagreement sweep ($k\ge2048$ with $d'=1024$, $k=4096$ with $d'=2048$),
the $k=4096$ drift-versus-$k$ cell ($d'=2048$), and the SmolLM2 student
($d'=960$) at $k=1024$. There $A$ fills $\R^{d'}$ and the per-position blind
subspace of \cref{thm:invariance} is empty.

\paragraph{Aggregation of the blind subspace over positions.} The intersection of
the per-position blind subspaces is $(\sum_i A_i)^\perp$, of dimension
$d'-\dim(\sum_i A_i) \ge d'-\sum_i \dim A_i$, with equality when the $A_i$ are in
general position. In the measurement reported in \cref{sec:discussion}
($V=20{,}000$), the dimension matches the general-position count except at
$N=8$ ($9$ rather than $8$, a one-dimensional overlap). It can collapse only once
$\sum_i \dim A_i \ge d'$, and here does so as soon as that holds, after about
$d'/(k-1)$ positions. The decomposition, by contrast, aggregates exactly. In a
separate random setup ($V=3000$, $d'=16$, $k=5$, one shared unembedding, $40$
positions with independent teacher distributions and student states), the sum
of \cref{eq:residual} over positions matches the summed full-vocabulary gradient
to within about $8\times10^{-15}$ in norm. Both checks use a dedicated script.

\paragraph{Off-policy: three-seed pilot.} Training Llama-3.2-1B from Llama-3.1-8B at three seeds,
with the cached arm at $8192$ bytes per position (the online arm stores
nothing): mean
training loss $0.34808$ (online) against $0.34791$ (cached), $p=0.75$; held-out
perplexity $11.655$ against $11.656$. The top-$256$ arm used $1536$ bytes per
position, a smaller and unmatched budget, and reached mean training loss
$0.3277$ and perplexity $12.114$; its training loss is lower than the cached
arm's but is computed over a $256$-token support rather than the full
vocabulary, so the two are not comparable and only the perplexity comparison is
informative. The paired test against top-$256$ gives $p=0.25$. Both are exact
sign-flip $p$-values, which at $n=3$ can only be $0.25$, $0.5$, $0.75$, or $1$.

\paragraph{Off-policy: detail on the six-seed comparisons.} The registered
prediction also named a downstream benchmark suite (MMLU, ARC-Challenge,
HellaSwag, GSM8K, HumanEval, IFEval), which was not run. Four of the eight
scoped arms were run: top-$k$ (ii, at $k=256$ and $1024$ only), feature matching (vi), online (vii) and the
cache (viii). Cached against online: 95\% bootstrap interval on the perplexity
difference $[0.0015, 0.0053]$ (the 90\% $t$ interval in \cref{sec:exp} is
$[0.0012,0.0055]$). The original three-seed figures
($p=0.75$, $p=0.25$, above) sat at or above the $n=3$ floor of $0.25$ and could
not have shown significance regardless of the effect. On Qwen3-1.7B/Qwen3-0.6B,
the first matched-storage teacher run, top-$1024$ wins at $1.5\times$ the
storage with Cohen's $d_z=40$; feature
matching there reaches $83{,}248$ perplexity despite a training loss that
visibly decreases (from $14.2$ to $0.75$ over $500$ steps): regressing the
student's hidden state onto the teacher's by mean-squared error trains, but does
not produce a model that predicts the next token competently. On
Gemma-3-4B/Gemma-3-1B, the only other teacher in \cref{tab:storage} with
$2d<6144$, the cache has both smaller storage and better quality, exactly the
frontier claim as registered. Qwen3-4B was chosen to isolate logit soft-capping
(present on Gemma-3-4B, absent on both Qwen3 teachers) at Gemma-3-4B's exact
$d=2560$; it favors top-$1024$, so soft-capping is not sufficient to explain the
split.

\paragraph{Off-policy: why the Qwen3 teachers favor top-$1024$.} Two
follow-ups, both on the matched-storage pairs and the same data stream. First,
an online full-vocabulary arm (teacher in the loop, no cache, otherwise
identical: learning rate $2\times10^{-5}$, $500$ steps, six seeds) on both Qwen3
pairs. Qwen3-1.7B: online $18.618$, cached $18.615$, top-$1024$ $18.336$; online
against cached $p=0.34$, $d_z=0.40$; online against top-$1024$ $p=0.03125$,
$d_z=31.6$. Qwen3-4B: online $13.881$, cached $13.885$, top-$1024$ $13.747$;
$p=0.31$, $d_z=-0.45$ and $p=0.03125$, $d_z=20.7$. The cache reproduces online
training, as Exactness requires, and the deficit belongs to full-vocabulary
supervision on these teachers. Second, the probability each teacher leaves
outside its own top-$k$ (\cref{tab:tailmass}), computed exactly from the full
distribution on $65{,}408$ training positions. Ranked by median tail mass beyond
top-$1024$, the teachers fall in nearly the order of the cache's advantage over
top-$1024$ ($+1.142$, $+0.038$, $+0.011$, $-0.279$, $-0.138$ perplexity for
Gemma-3-4B, Llama-3.2-3B, SmolLM2-1.7B, Qwen3-1.7B, Qwen3-4B); only the two Qwen3
teachers swap, and the mean tail mass and the entropy give the same ranking.
Where the top $1024$ tokens hold all but $0.04$--$0.06\%$ of a typical
position's mass, top-$1024$ supervision is nearly exact, which leaves the cache
little to add; why full-vocabulary supervision then does slightly worse is not
settled here.

\begin{table}[h]
\centering
\caption{Teacher probability outside its own top-$k$, on the matched-storage
training stream ($65{,}408$ positions), with mean entropy. ``Cache gain'' is
top-$1024$ minus cache perplexity from \cref{tab:matched}.}
\label{tab:tailmass}
\begin{tabular}{l r r r r r}
\toprule
Teacher & \multicolumn{2}{c}{Tail beyond top-$1024$} & Tail beyond top-$64$ & Entropy & Cache gain \\
\cmidrule(lr){2-3}
 & median & mean & mean & (nats) & (ppl) \\
\midrule
Gemma-3-4B   & 0.0121 & 0.0201 & 0.151 & 3.59 & $+1.142$ \\
Llama-3.2-3B & 0.0024 & 0.0097 & 0.064 & 2.28 & $+0.038$ \\
SmolLM2-1.7B & 0.0020 & 0.0087 & 0.061 & 2.20 & $+0.011$ \\
Qwen3-1.7B   & 0.0006 & 0.0070 & 0.036 & 1.68 & $-0.279$ \\
Qwen3-4B     & 0.0004 & 0.0055 & 0.028 & 1.53 & $-0.138$ \\
\bottomrule
\end{tabular}
\end{table}

\paragraph{Off-policy: the five-teacher matched-storage comparison, in
full.} Qwen3-4B/Qwen3-1.7B, six seeds, matched at top-$1024$'s $6144$
bytes: cached reached $13.8852$ perplexity (bootstrap CI
$[13.8818,13.8892]$) against top-$1024$'s $13.7473$ ($[13.7447,13.7496]$),
$p=0.03125$, Cohen's $d_z=25.2$; feature matching reached $117{,}018$
perplexity, $p=0.03125$ against cached. Llama-3.2-3B/Llama-3.2-1B ($d=3072$), six
seeds, both arms at $6144$ bytes: cached $11.680$ against top-$1024$'s
$11.718$, $p=0.03125$, $d_z=-6.8$; feature matching $55{,}517$.
SmolLM2-1.7B/SmolLM2-360M, six seeds, cached at $4096$ bytes against
top-$1024$ at $6144$: cached $12.558$ against $12.569$, $p=0.03125$,
$d_z=-2.4$; feature matching $684.7$. Because this student has $d'=960<k-1$,
its top-$1024$ arm has no per-position blind subspace, so this pair tests a
different regime from the other four. Unembedding geometry for all five
teachers, computed from the unembedding matrix $U$ directly (condition
number $\sigma_{\max}/\sigma_{\min}$ on $50{,}000$ sampled rows, or all
$49{,}152$ rows for SmolLM2, as in \cref{tab:e2}, and anisotropy, the norm of
the mean row divided by the mean row norm): Qwen3-4B $59.5$, $0.285$;
Qwen3-1.7B $33.7$, $0.226$; Gemma-3-4B $30.0$, $0.166$; Llama-3.2-3B $60.8$,
$0.371$; SmolLM2-1.7B $115.2$, $0.270$. On the first three teachers both
quantities were ordered exactly as the win/loss split. The two added
teachers break that ordering: both have higher condition number than
either Qwen3 teacher and higher anisotropy than Qwen3-1.7B, yet the cache
wins on both. Neither quantity predicts the outcome.

\paragraph{On-policy: three-seed estimate.} At three seeds the high-entropy gap
was $74\times$ the low-entropy one; six seeds give $104\times$ (\cref{sec:exp}).

\paragraph{On-policy: exploratory student-selection data point.} Student
selection is covered by \cref{thm:residual,thm:invariance} only within regions
where the set does not change. Student-selected top-$k$ gives mean held-out KL
loss $0.4224$ (low entropy) and $3.5231$ (high entropy), against $0.4513$ and
$4.0296$ for teacher-selected top-$k$ (\cref{sec:exp}). No exact-supervision row exists for
student selection, so no gap is reported for that condition.

\paragraph{Mechanism at $\alpha_{ce}=1$.} Same teacher, student, $k=256$, data and
six seeds as the $\alpha_{ce}=0$ run, with loss $\mathcal{L}_\Sset$ (or
$\mathcal{L}$) plus one unit of cross-entropy. Mean student tail mass at steps
$0,50,250,500$: top-$k$ $0.0496\to0.0525\to0.0544\to0.0542$, full vocabulary
$0.0496\to0.0468\to0.0471\to0.0468$ (teacher $0.0216$; seed standard deviation
at most $2\times10^{-5}$). Mean $|m_p-m_q|$ rises from $0.0280$ to $0.0326$ for
top-$k$ and falls to $0.0252$ for the control. The final gap, $0.0074$, is
$4.25\times$ smaller than at $\alpha_{ce}=0$ ($0.0313$). Final expected
calibration error is $0.019$ (top-$k$) against $0.013$ (full vocabulary).

\paragraph{Mechanism: determinism detail.} Checkpoints at steps
0, 50, 250, and 500. An earlier three-seed run (seeds $0$--$2$) used
the identical configuration, so its seeds rerun three of the six: across those
$24$ arm-seed-checkpoint cells, absolute tail
mass differs by at most $7\times10^{-5}$ (mean $2\times10^{-5}$), more than two
orders of magnitude below the effects discussed. The qualitative conclusions
are therefore not sensitive to run-to-run non-determinism, which seed
variance (six seeds, \cref{sec:exp}) does not measure.

\paragraph{Mechanism: which tokens absorb the tail mass.} At $\alpha_{ce}=0$,
step $500$ of the same run reported in \cref{fig:teaser}b
(Llama-3.1-8B~$\to$~Llama-3.2-1B, $k=256$), we accumulate the student's tail
probability mass ($V\setminus\Sset$ under the student's own distribution) per
token id across every supervised position in the probe, rather than only the
aggregate scalar $m_p$. The tokens carrying the most absorbed mass are not
rare or exotic: they are short, extremely common subword fragments that begin
many different words: single-letter or short byte-pair prefixes such as
``re'', ``p'', ``f'', ``s'', ``pre'', ``h'', ``t'', ``b'', ``c'', ``m'', ``de'',
``g'', ``un'', ``co'', alongside a handful of short common words (``free'',
``first'', ``long'', ``over'', ``up'', ``work'', ``out'', ``state'', ``well'',
``self'', ``back''). No rare tokens, punctuation, or non-English script
appears near the top. This is consistent with the mechanism
\cref{thm:invariance} describes. These fragments are common enough to carry
real probability mass at many positions, but rarely the single most likely
continuation, so they are exactly the class of token most likely to sit just
outside a $k{=}256$ cutoff at any given position, and therefore in the blind
subspace the truncated objective's gradient cannot see. They are not a
semantically distinctive class the model is specifically failing to
represent.

\paragraph{Disagreement: limitation detail.} Because the measured best
$k^\star=1024$ sits at the edge of the swept range ($k\in\{8,64,256,1024\}$) in
all three settings, held-out $\KL$ was still improving at the largest $k$
tested. With $\KL$ monotone over $\{8,64,256,1024\}$, $k^\star$ was forced to the
largest swept value unless the effect reversed inside the range, so the original
design could not have detected an optimum beyond its largest $k$: weak evidence
against the hypothesis and strong evidence that the sweep was too narrow, which
the widened six-seed sweep in \cref{sec:exp} then checked. Widened to six seeds at $\{2048,4096\}$, each doubling of $k$ lowers
$\KL$ by $0.573\%$ then $0.280\%$ (bild-style), $0.339\%$ then $0.121\%$
(pretrain-style), and $0.226\%$ then $0.136\%$ (toolcall-style). Holding capacity
fixed: two of the three settings, bild-style and toolcall-style, use the same
pair as one of Capacity's points (Qwen3-8B$\to$Qwen3-0.6B), differing only in
corpus. Both fall monotonically in $k$ with no interior optimum
($0.967\to0.885\to0.858$ and $0.508\to0.496\to0.491$ at $k=64,256,1024$), the
same pattern Capacity's generic-corpus run gives this pair ($0.825$ at $k=64$):
corpus shifts the absolute level but not whether an optimum exists. This is
consistent with capacity governing the qualitative $k$-behavior and corpus its
magnitude, though two corpora at one pair is not enough to call this settled.

\paragraph{Precision: synthetic versus real.} On a synthetic teacher
($d=4096$), int8 caching at $4100$ bytes per position gives
$\KL=2.60\times10^{-11}$ and maximum log-probability error $0.0887$. On the real
teacher (Qwen3-1.7B, $d=2048$, so $2052$ bytes at int8), the cache again sits
between top-$256$ and top-$1024$ in size, but the error is $\KL=2.02\times10^{-4}$, about $7.8\times10^{6}$ times (nearly seven orders of magnitude) larger, with a
comparable maximum log-probability error ($0.0943$). The real-teacher number is
authoritative; the synthetic figure shows how badly random geometry understates
real quantization error. The two teachers differ in $d$, so the comparison
is between geometries, not matched configurations.

\paragraph{Capacity: remaining detail.} Beyond the values in \cref{sec:exp},
Qwen3-8B$\to$Qwen3-0.6B reaches $0.753$ at $k=256$, and all three pairs fall
monotonically in $k$. A fourth configuration with student equal to teacher gives
$0.0013$ at $k=64$ and would inflate the pair range to a factor of $654$; we
exclude it from the $1.92$, since a self-distillation point measures nothing
about capacity. Because Qwen3-32B was not run, we do not extend the conclusion
to the largest teacher scale scoped.

\section{Design tests: full results}
\label{app:torun}

The design specified four tests (the blind-subspace probe, drift versus $k$,
student-selected retained sets, and the residual plateau) precisely enough
to execute, each with its research question and, where one was stated, its prediction, before any of
them had been run. All four have since been run; their full results follow.

\paragraph{Blind-subspace probe (tests \cref{thm:invariance} directly).
Confirmed; see \cref{sec:exp}.} Question: is $\nabla_h\mathcal{L}_\Sset$
exactly zero on $A^\perp$, and is $\nabla_h m_p$ not? Method: at any
checkpoint, and for each of a few hundred supervised positions, form
$A = \mathrm{span}\{w_v - w_{v_0}: v\in\Sset\}$ by a thin QR of the
$(k-1)\times d'$ matrix of centered head rows, let $P$ project onto $A^\perp$,
and report $\|P\nabla_h\mathcal{L}_\Sset\|/\|\nabla_h\mathcal{L}_\Sset\|$ and
$\|P\nabla_h m_p\|/\|\nabla_h m_p\|$. Prediction: the first is at float
precision zero; the second is bounded away from zero and decreases toward zero
as $k \to V$. Cost: one backward pass per position; no training. A
prototype on synthetic data
gave $10^{-16}$ and $0.55$--$0.73$ respectively at $k \ll d'$, falling to
$0.02$ at $k = d'$; run on Qwen3-8B/Qwen3-0.6B at $256$ real positions, the
prediction held exactly: median ratio $2$--$3\times10^{-15}$ for the first
quantity at every $k$, and $0.32 \to 0.17 \to 0.0027 \to 0$ for the second as
$k$ ran $64 \to 256 \to 1024 \to V$.

\paragraph{Drift versus $k$, with the negative control
(tests \cref{thm:invariance} quantitatively). Confirmed; see the
Drift-versus-$k$ paragraph in \cref{sec:exp}.}
Method detail not given there: drift is $|m_p^{\text{final}} - m_q|$, with
$m_q$ the teacher's tail mass outside the arm's own retained set, so part of
the fall comes from $m_q$ itself ($0.0216$ at $k=256$, $0.0018$ at $k=4096$).
The full-vocabulary control has no retained set of its own; its drift
($0.0255$) is measured against the $k=256$ set, so it is comparable with the
top-$256$ cell ($0.057$), not with $k=4096$.

\paragraph{Student-selected retained sets (probes the scope caveat).}
Question: does the blind subspace survive when $\Sset$ depends on $h$, as
in \citet{sfd2026}? Method: repeat Mechanism with $\Sset$ the union of
student and teacher top-$10$, recomputed each step, six seeds. Prediction: not
stated; we do not know, and the theorems cover this case only within regions
where the set does not change. Result: the
final drift is smaller under union selection than under fixed
teacher-selection at the same $k=10$ ($0.200$ against $0.327$, $p=0.03125$,
Cohen's $d_z=884$): the student-dependent case drifts less, not more, in this
configuration. Why it matters: this is
the one gap between our result and the setting of the claim we correct; the
theorems say nothing across changes of the set, and this single comparison is
not enough to close that gap, but it does not point toward alarm either.

\paragraph{Residual plateau (tests \cref{thm:residual}'s consequence).
Falsified but mechanistically explained; see the Residual-plateau
paragraph in \cref{sec:exp}.}
Registered prediction: $\|\nabla\mathcal{L}_\Sset\|$ goes to zero while
$\|\nabla\mathcal{L}\|$ plateaus; neither held. The lower bound's tightness
(its ratio to the measured residual, averaged over positions) stays at or below
$0.42$; the ratio of the averaged quantities reaches $0.46$ at the end of the run. Further detail: the bound's $\approx40\%$ recovery of the
residual's magnitude is consistent with the $29$--$37\%$ found in
\cref{app:numerics}'s independent synthetic check, and its rise over the
run ($38\%$) outpaces the residual's own rise ($14\%$) over the same
window, which is why the bound's tightness itself increases rather than
staying fixed. After its first $250$ steps the residual rises with small fluctuations;
the correlation between bound and residual is $r=0.79$ over all checkpoints and
$0.9995$ after step $250$. Corollary~\ref{cor:lowerbound} does not predict a flat
residual here, only one bounded away from zero by a quantity this trajectory
grows rather than holds fixed. We report the falsification as registered rather
than reweight it after the fact.

\end{document}

%% file: affiliations.tex
\DeclareAffiliation{hbku}{%
  College of Science and Engineering, Hamad Bin Khalifa University, Doha, Qatar}

\DeclareAffiliation{tamu}{%
  Department of Computer and Electrical Engineering,
  Texas A\&M University, College Station, TX, USA}

\DeclareAffiliation{iub}{%
  Luddy School of Informatics, Computing, and Engineering,
  Indiana University Bloomington, Bloomington, IN, USA}
